\pdfoutput=1
\documentclass[11pt]{article}

\usepackage[final]{acl}

\usepackage{times}
\usepackage{latexsym}

\usepackage{amsmath}
\usepackage{diagbox}
\usepackage{caption}
\usepackage{amsfonts,amssymb}
\usepackage{subfigure}
\usepackage{graphicx}
\usepackage{bm}
\usepackage{multirow}
\usepackage[ruled,lined,commentsnumbered]{algorithm2e}
\usepackage{hyperref}
\usepackage{booktabs}
\usepackage{bbm}
\usepackage[inkscapelatex=false]{svg}
\usepackage{pifont}% http://ctan.org/pkg/pifont
\usepackage{stfloats}
\usepackage{multicol}
\usepackage{float}
\usepackage{array}
\newcommand{\PreserveBackslash}[1]{\let\temp=\\#1\let\\=\temp}
\newcolumntype{C}[1]{>{\PreserveBackslash\centering}p{#1}}
\newcolumntype{R}[1]{>{\PreserveBackslash\raggedleft}p{#1}}
\newcolumntype{L}[1]{>{\PreserveBackslash\raggedright}p{#1}}

\usepackage{mathrsfs}

\usepackage[T1]{fontenc}
\usepackage[utf8]{inputenc}

\usepackage{microtype}

\usepackage{inconsolata}

\title{Decoder-only Streaming Transformer for Simultaneous Translation}

\author{
    Shoutao Guo \textsuperscript{\rm 1,3},
    Shaolei Zhang \textsuperscript{\rm 1,3},
    Yang Feng \textsuperscript{\rm 1,2,3}\thanks{ \ \ Corresponding author: Yang Feng.} \\
        \textsuperscript{\rm 1}{Key Laboratory of Intelligent Information Processing,} \\ Institute of Computing Technology, Chinese Academy of Sciences (ICT/CAS) \\
    { \textsuperscript{\rm 2} {Key Laboratory of AI Safety, Chinese Academy of Sciences}} \\
    { \textsuperscript{\rm 3} {University of Chinese Academy of Sciences, Beijing, China}} \\
     \texttt{\href{mailto:guoshoutao22z@ict.ac.cn}{guoshoutao22z@ict.ac.cn}, \href{mailto:zhangshaolei20z@ict.ac.cn}{zhangshaolei20z@ict.ac.cn}, \href{mailto:fengyang@ict.ac.cn}{fengyang@ict.ac.cn}}  }

\begin{document}
\maketitle
\begin{abstract}
Simultaneous Machine Translation (SiMT) generates translation while reading source tokens, essentially producing the target prefix based on the source prefix. To achieve good performance, it leverages the relationship between source and target prefixes to exact a policy to guide the generation of translations. Although existing SiMT methods primarily focus on the Encoder-Decoder architecture, we explore the potential of Decoder-only architecture, owing to its superior performance in various tasks and its inherent compatibility with SiMT. However, directly applying the Decoder-only architecture to SiMT poses challenges in terms of training and inference.
To alleviate the above problems, we propose the first Decoder-only SiMT model, named Decoder-only Streaming Transformer (DST). Specifically, DST separately encodes the positions of the source and target prefixes, ensuring that the position of the target prefix remains unaffected by the expansion of the source prefix. Furthermore, we propose a Streaming Self-Attention (SSA) mechanism tailored for the Decoder-only architecture. It is capable of obtaining translation policy by assessing the sufficiency of input source information and integrating with the soft-attention mechanism to generate translations. Experiments demonstrate that our approach achieves state-of-the-art performance on three translation tasks\footnote{Code is at \url{https://github.com/ictnlp/DST}}.

%On one hand, the continuous arrival of source tokens causes a dynamic shift in positions of generated translation, which necessitates re-encoding of translation and results in a increase in inference cost. On the other hand, training the model based on prefix pairs escalates the training cost of a sentence pair significantly. 
\end{abstract}

\section{Introduction}
Simultaneous Machine Translation (SiMT) \citep{reinforcement, DBLP:conf/acl/MaHXZLZZHLLWW19} is designed for generating translations in real-time scenarios such as online conferences and real-time subtitles. It predicts the target tokens (i.e., target prefix) based on the already read source tokens (i.e., source prefix), aiming to achieve good tradeoffs between latency and translation quality. During training, SiMT models need to learn the correspondence between source and target prefixes, crucial for extracting policies that ensure superior performance during inference \citep{DualPath}.

Existing research on SiMT primarily focuses on the Encoder-Decoder architecture and is categorized into fixed and adaptive policies. For fixed policy \citep{dalvi-etal-2018-incremental, DBLP:conf/acl/MaHXZLZZHLLWW19, multiPath}, the model utilizes heuristic rules to determine the source prefix used for generating translations, which ignores the correspondence between the source and target prefixes. This may lead to redundant or missing source information during translation, resulting in inferior performance \citep{gaussian}. For adaptive policy \citep{DBLP:conf/iclr/MaPCPG20}, the model dynamically decides whether to read or output tokens based on the relationship between the source and target prefixes. This dynamic adjustment of policy in response to the translation status allows for improved tradeoffs \citep{zhao-etal-2023-adaptive}. However, there is a lack of exploration in SiMT regarding the Decoder-only architecture.

\begin{figure}[t]
\centering
\subfigure[Encoder-Decoder Architecture.]{
\includegraphics[width=3.0in]{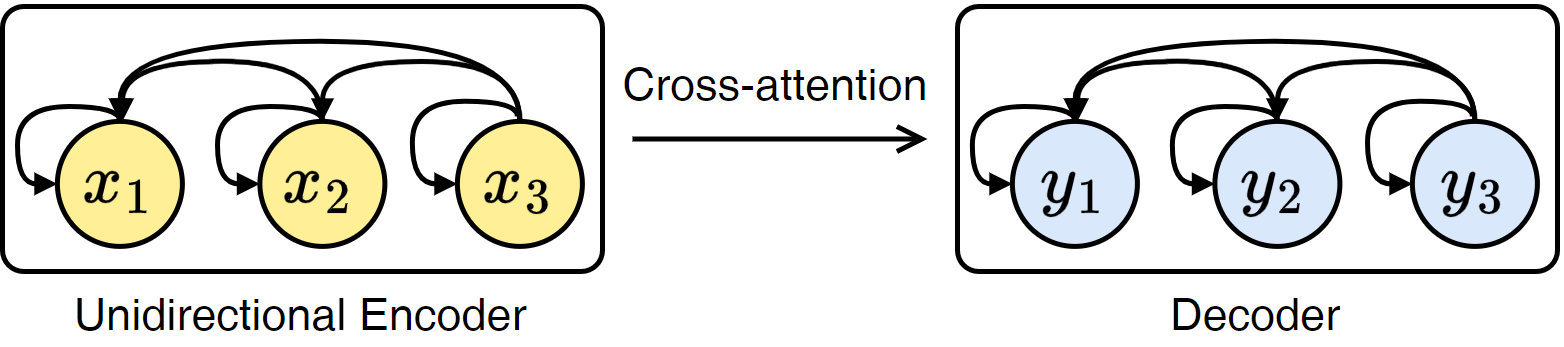}
}\hspace{-0.3cm}
\subfigure[Decoder-only Architecture.]{
\includegraphics[width=2.35in]{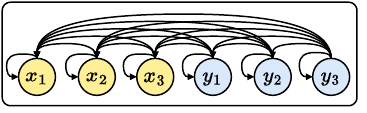}
}\hspace{-0.3cm}

\caption{Comparison of Encoder-Decoder architecture and Decoder-only architecture.}
\label{fig1:cpmpar}
\end{figure}

With the rise of language models, the Decoder-only architecture has exhibited superior performance across diverse tasks \citep{touvron2023llama, gemmateam2024gemma}. As illustrated in Figure \ref{fig1:cpmpar}, the Decoder-only architecture, compared to the Encoder-Decoder architecture with an equivalent number of parameters, can support more layers, thereby offering parameter efficiency and better scalability \citep{liu-etal-2023-enhancing-scalability}. Importantly, SiMT relies on unidirectional encoding \citep{multiPath}, and the Decoder-only architecture seamlessly accommodates this requirement. Therefore, we explore the capability of Decoder-only architecture in SiMT.

However, directly applying the Decoder-only architecture to the SiMT task poses challenges in both training and inference. During inference, with the arrival of each source token, there is a position increase of the generated target prefix, necessitating the model to re-encode the target prefix. This exacerbates the inference cost, particularly at low latency \citep{wang2024conversational}. During training, the model learns to predict the corresponding target prefix based on a given source prefix. Consequently, it is necessary to construct corresponding target prefixes for all possible source prefixes in a sentence pair. This limitation hinders the model from learning the translation policy and leads to an increase in training costs compared to the Encoder-Decoder architecture.

To overcome the above limitations, we propose the first SiMT model based on Decoder-only architecture, named the \textbf{D}ecoder-only \textbf{S}treaming \textbf{T}ransformer (DST). To alleviate the issue of re-encoding, DST encodes the positional information of the source prefix and the target prefix separately. This ensures that the expansion of the source prefix does not impact the position of the generated target prefix, thereby reducing the inference costs. To assess the contribution of partial source information to generating target tokens, DST uses the proposed Streaming Self-Attention (SSA) in replace of the conventional masked self-attention in the Decoder layer to decrease training costs and derive a translation policy.

During training, SSA can consider all possible source prefixes for the target prefixes in a sentence pair. Specifically, SSA predicts attention allocation for different source prefixes and combines it with the soft-attention mechanism to obtain expected attention for all source tokens and tokens in the target prefix. This expected attention is then utilized to derive the context vector.
By leveraging SSA, the model learns the importance of all source prefixes in translating the target prefix, thereby reducing training costs. During inference, SSA accumulates the allocated attention from all prefixes of the input source tokens, enabling an assessment of the sufficiency of input source information for generating translation. The model utilizes this assessment to determine whether to read or generate tokens, thereby acquiring the translation policy.
Experiments demonstrate that DST achieves state-of-the-art performance on three tasks.

%Furthermore, DST introduces an adaptive policy tailored for the Decoder-only architecture. This policy is derived from a crucial property in the Decoder-only architecture, where target tokens concurrently attend to both source and previous target tokens \citep{zhang2022examining}. If the model allocates less attention to the source information, it prioritizes language modeling capability over translation fidelity. In such cases, the source prefix may lack the necessary content for the current translation, prompting the model to continue reading the remaining tokens \citep{ITST}. Alternatively, the model should persist in generating translation. To learn this policy, we propose a novel expectation training method, which aids the model in determining the appropriate allocation of source attention with a source prefix and target prefix.
%During this process, the model can implicitly learn the possible correspondences between all source and target prefixes in a sentence pair, significantly reducing training costs. 

\section{Background}
\paragraph{Simultaneous Machine Translation}
The SiMT model \citep{DBLP:conf/acl/MaHXZLZZHLLWW19} dynamically reads the source sentence $\mathbf{x}$ = $(x_1, ..., x_J)$ with length $J$ and generates translation $\mathbf{y}$ = $(y_1, ..., y_I)$ with length $I$ based on a policy. To articulate the policy, we introduce $g_i$, representing the number of source tokens involved in translating $y_i$. Thus, the translation policy from $\mathbf{x}$ to $\mathbf{y}$ can be denoted as $\mathbf{g}$ = $(g_1, ..., g_I)$. During training, the SiMT model undergoes optimization by minimizing the cross-entropy loss:
\begin{equation}
\mathcal{L}_{simt} = - \sum\limits_{i = 1}^{I} \log p(y^{\star}_i \mid \mathbf{x}_{\leq g_i}, \mathbf{y}_{<i}),
\end{equation}
where $y^{\star}_i$ represents the ground-truth token.
\paragraph{Masked Self-Attention}
Masked self-attention allows each position in the decoder to attend to all positions in the decoder up to and including that position \citep{DBLP:conf/nips/VaswaniSPUJGKP17}, ensuring the auto-regressive generation. Given the target hidden states $\mathbf{s}$ = $(s_1, ..., s_I)$, the context vector is computed as follows:
\begin{equation}
e_{i,k} = s_i W^Q (s_k W^K)^{\intercal} / \sqrt{d},
\label{scaledDot}
\end{equation}
\begin{equation}
  \alpha_{i,k} = \left\{\begin{matrix}  
  {\exp (e_{i,k})} / {\sum\limits_{l=1}^{i} \exp (e_{i,l})}  & \text{if} \;\; k \leq i  \\[0.05cm]
  0 &\text{otherwise}
  \end{matrix}\right. ,
\label{assign_prob}
\end{equation}
\begin{equation}
c_{i} = \sum_{k=1}^{i}\alpha_{i,k}(s_kW^V),
\end{equation}
where $W^Q$, $W^K$ and $W^V$ are projection parameters, and $d$ denotes the dimension of inputs.

\section{Method}
\label{all_method}
In this section, we introduce the Decoder-only Streaming Transformer (DST). DST adopts the Decoder-only architecture and employs the proposed Streaming Self-Attention (SSA) in place of masked self-attention at each layer. During inference, SSA accumulates the attention assigned to the input source tokens to assess the sufficiency of source information and obtain the translation policy accordingly. During training, DST leverages our proposed constraints to ensure the learning of the SSA mechanism. The details of DST are introduced as follows.

\begin{figure}[t]
    \centering
    \includegraphics[width=2.3in]{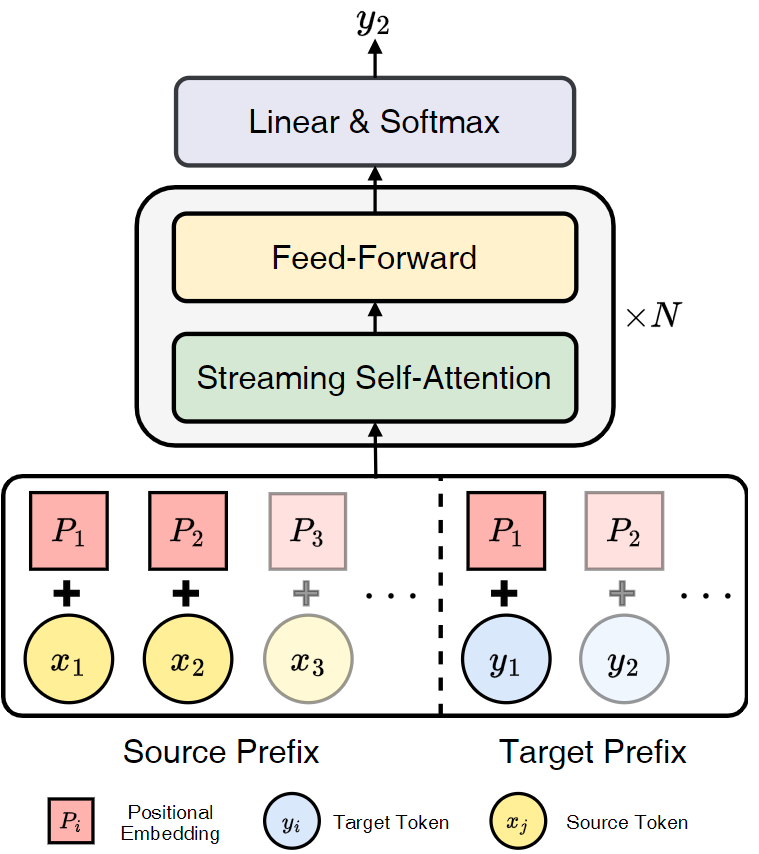}
    \caption{The architecture of DST. It shows the moment when DST generates $y_2$ after reading two source tokens.}
    \label{fig-model}
\end{figure}

\subsection{Model Architecture}
We present the model architecture of DST in Figure \ref{fig-model}. DST takes the concatenation of the source and target prefixes as input, encoding the positional information of both the source and target prefixes separately. In this way, the expansion of the source prefix does not affect the position of the generated target prefix, preventing the re-encoding of the generated target prefix. In each layer, DST replaces the masked self-attention module \citep{DBLP:conf/nips/VaswaniSPUJGKP17} in the conventional Decoder-only architecture with our proposed Streaming Self-Attention (SSA) module. During inference, the SSA module will determine the policy and integrate with the soft-attention mechanism to derive the context vector and predict the next token.

\subsection{Streaming Self-Attention}
\label{SSA}
In order to address the SiMT task, DST incorporates Streaming Self-Attention (SSA) in place of masked self-attention. By concatenating the source sentence and the target sentence as input, SSA adopts masked self-attention within the source sentence \citep{DBLP:conf/nips/VaswaniSPUJGKP17}. For each target token, SSA initially allocates its attention probability to all possible source prefixes and subsequently combines the allocated attention with the soft-attention mechanism to compute the expected attention probability for source tokens and preceding target tokens. The context vectors of target tokens are computed by utilizing the expected attention. We then provide a detailed explanation of the SSA mechanism as applied to target tokens during training.

Given the source sentence $\mathbf{x}$ and the target sentence $\mathbf{y}$, the hidden state sequence input to SSA is denoted as $\mathbf{h}$ = $(h^s_1, ..., h^s_J, h^t_1, ..., h^t_I)$, where $h^s_j$ denotes the source hidden state and $h^t_i$ represents the target hidden state. To articulate the allocated attention to the source prefixes, we define $p_{i,j}$, which signifies the allocated attention of the target token $y_i$ to the source prefix $\mathbf{x}_{\leq j}$. The calculation of $p_{i,j}$ is computed by utilizing the relationship between the source and target prefixes:
\begin{equation}
p_{i,j} = \mathrm{sigmoid}(\frac{\overline{h}_i^tU^Q(\overline{h}_j^sU^K)}{\sqrt{d}}),
\end{equation}
where $U^Q$ and $U^K$ are learnable parameters, and $d$ denotes the dimension of hidden state. $\overline{h}_i^t$ and $\overline{h}_j^s$ represents the mean pooling of hidden states corresponding to the target prefix $\mathbf{y}_{\leq i}$ and source prefix $\mathbf{x}_{\leq j}$, respectively. Subsequently, we utilize $p_{i,j}$ to obtain the expected attention of each target token towards both source tokens and preceding target tokens.

Taking into account all possible source prefixes when translating the target token $y_i$, the expected attention of target token $y_i$ to the source token $x_j$ is calculated as:
\begin{equation}
\alpha^{s}_{i,j} = \sum\limits_{m=j}^{J}\frac{p_{i,m}\exp(e^s_{i,j})}{\sum_{l=1}^{m}\exp{(e^{s}_{i,l})}},
\end{equation}
where $e^s_{i,j}$ denotes the scaled dot-product of $h^t_{i}$ and $h^s_{j}$, as illustrated in Eq.(\ref{scaledDot}). The first summation considers all possible source prefixes that include $x_j$, and the term inside summation reflects the attention probability associated with $p_{i,m}$ and soft-attention. Correspondingly, the expected attention of $y_i$ to the target token $y_k$$ (k \in [1, i])$  is:
\begin{equation}
\alpha^{t}_{i,k} = (1 - \sum\limits_{j=1}^{J}p_{i,j})\frac{\exp({e^t_{i,k}})}{\sum_{l=1}^{i}\exp({e^t_{i,l}})},
\end{equation}
where $\sum_{j=1}^{J}p_{i,j}$ represents the total attention to the source tokens and $e^t_{i,k}$ is the scaled dot-product of $h_i^t$ and $h_k^t$. Therefore, we can get the context vector of $y_i$ by considering both source tokens and previous target tokens:
\begin{equation}
c^{t}_i = \sum\limits_{j=1}^J \alpha_{i,j}^s(h_j^sW^V) +  \sum\limits_{k=1}^i \alpha_{i,k}^t(h_k^tW^V),
\end{equation}
where $W^{V}$ denotes the projection parameters.
%We subsequently calculate the expected attention of $y_i$ on the previous target tokens. Considering the meaning of $p_{i,j}$, $\sum_{j=1}^{J}p_{i,j}$ represents the expected attention to source tokens when the SiMT model translate generate $y_i$.

%Therefore, SSA allocates attention to source prefixes based on their correspondence with target prefixes. It integrates the allocated attention with the soft-attention to obtain the context vector. 
The above is the operational mechanism of SSA during training. During inference, SSA utilizes the input source tokens and the previously generated target tokens to generate the context vector.

\subsection{Inference}
After introducing SSA, DST employs it to derive a policy, which instructs the model to generate translation. The SiMT policy generally assesses the adequacy of the source information to determine whether to proceed with generating the translation or to read additional source tokens \citep{zhang2023unified}. In DST, SSA initially allocates attention to different source prefixes using $p_{i,j}$, subsequently integrating the allocated attention with the soft-attention. Consequently, $\sum_{j=1}^{m} p_{i,j}$ signifies the accumulated attention to the current available source prefix $\mathbf{x}_{\leq m}$ and quantifies the sufficiency of source information. By setting an appropriate threshold $\delta_{infer}$, DST deems the source information sufficient under the condition:
\begin{equation}
    \sum_{j=1}^{m} p_{i,j} > \delta_{infer},
\end{equation}
where $m$ is the number of input source tokens. At this point, DST is capable of generating the translation based on the existing source tokens. Otherwise, the model should continue reading the source tokens until the aforementioned condition is met or the entire sentence is input to our model. It is noteworthy that DST will generate the translation only when most layers meet the conditions.

\subsection{Training Method}
\label{training_method}
To facilitate the learning of SSA mechanism, we introduce the training method. The essence lies in the learning of $p_{i,j}$. Although we may not directly provide an optimization objective for $p_{i,j}$, we can leverage the characteristics of SiMT tasks \citep{zhang-etal-2022-wait} and the properties of $p_{i,j}$ to induce its learning.
Therefore, in addition to the cross-entropy loss $\mathcal{L}_{simt}$, we propose three additional constraints and a curriculum learning strategy to aid in the training of DST. 
\paragraph{Summation Constraint}
Based on the inherent property of $p_{i,j}$, we introduce the summation constraint. As described in the section \ref{SSA}, $p_{i,j}$ represents the attention probability allocated to the source prefix $x_{\leq j}$ and $\sum_{j=1}^{J} p_{i,j}$ reflects the total attention on all source tokens. Therefore, it should be equivalent to the sum of attention to source tokens in masked self-attention \citep{DBLP:conf/nips/VaswaniSPUJGKP17}:
\begin{equation}
    \mathcal{L}_{sum} = \sum\limits_{i=1}^{I} \|\sum_{j=1}^{J} p_{i,j} - \beta_i\|_2 ,
\end{equation}
where $\beta_i$ denotes the attention to the source tokens in masked soft-attention. It is computed as:
\begin{equation}
    \beta_i = \frac{\sum_{j=1}^{J}\exp({e^s_{i,j}})}{\sum_{j=1}^{J}\exp({e^s_{i,j}}) + \sum_{k=1}^{i}\exp({e^t_{i,k}})},
\end{equation}
where $e^s_{i,j}$ is the scaled dot-product of target token $y_i$ and source token $x_j$, and $e^t_{i,k}$ represents the scaled dot-product between the target token $y_i$ and its preceding target token $y_k$.
\paragraph{Latency Constraint}
In addition to the summation constraint, we introduce a latency constraint for DST. Without the latency constraint, SSA tends to allocate excessive attention to longer source prefixes during training \citep{DBLP:conf/iclr/MaPCPG20}. This tendency encourages the model to read more source tokens during inference, resulting in high latency. According to \citet{ITST}, the alignment favoring low latency tends to concentrate near the diagonal between source and target sentences. Expressing the attention allocation for source prefixes in sentence pair ($\mathbf{x}$, $\mathbf{y}$) as $\mathbf{p} = (p_{i,j})_{I\times J}$, the latency constraint aims to encourage SSA to allocate more attention to source prefixes closer to the diagonal. To make it, we introduce the cost matrix $\mathbf{C}=(C_{i,j})_{I\times J}$, where $C_{i,j}$ is computed as:
\begin{equation}
    C_{i,j} = \frac{1}{\max(I, J)} \max(|j-i\times \frac{J}{I}| - \epsilon, 0),
\end{equation}
where $|j-i\times \frac{J}{I}|$ quantifies the degree of deviation from the diagonal between the target token $y_i$ and the source prefix $\mathbf{x}_{\leq j}$. Therefore, $C_{i,j}$ supports attention distributions that are close to the diagonal. The hyperparameter $\epsilon$ controls the tolerance level for the deviation. We then obtain the latency constraint as:
\begin{equation}
    \mathcal{L}_{lat} = \sum\limits_{i=1}^{I} \sum\limits_{j=1}^{J} p_{i,j}C_{i,j}.
\end{equation}
Further explanation about latency constraint is provided in the Appendix \ref{sec:latency_constraint}.

\paragraph{Consistency Constraint}
During training, DST utilizes summation and latency constraints to ensure the learning of $p_{i,j}$ in each layer. However, there are variations among different layers \citep{yang-etal-2020-sub}, and our policy is obtained by integrating decisions from different layers. Without constraints on the decisions between different layers, the presence of outlier layers may result in excessively high latency for the learned policy. Therefore, we propose a consistency constraint to ensure consistency across decisions at different layers. Denoting the $p_{i,j}$ in the $n$-th layer as $p^{(n)}_{i,j}$, and the consistency constraint is calculated as follows:
\begin{align}
    \mathcal{L}_{con} = \sum\limits_{i=1}^{I}\sum\limits_{j=1}^{J}\sum\limits_{n=1}^{N} \frac{1}{N}\|p^{(n)}_{i,j} - \overline{p}_{i,j} \|_2,
\end{align}
where $N$ is the number of layers and $\overline{p}_{i,j}$ is computed as follows:
\begin{equation}
    \overline{p}_{i,j} = \frac{1}{N}\sum\limits_{n=1}^{N}p^{(n)}_{i,j}.
\end{equation}

Therefore, we can get the overall training objective of DST:
\begin{equation}
    \mathcal{L} = \mathcal{L}_{simt} + \mathcal{L}_{sum} + \mathcal{L}_{lat} + \mathcal{L}_{con}.
\end{equation}
\paragraph{Curriculum Learning Strategy}
If we train DST based on the training objective $\mathcal{L}$ directly, the model learns to generate translation using the entire source sentence. However, DST generates translations based on source prefixes during inference, leading to a discrepancy between training and inference \citep{zhang-feng-2022-reducing}. To mitigate this problem, we train DST to generate translations based on the source prefix by masking out subsequent source tokens during training. Specifically, by setting the threshold $\delta_{train}$, DST masks out the source tokens after the shortest prefix $\mathbf{x}_{\leq m}$ that satisfies the condition $\sum_{j=1}^{m} p_{i,j} > \delta_{train}$. Additionally, we introduce curriculum learning \citep{bengio2009curriculum}, which gradually transitions the available source information from the entire sentence to the prefix consistent with the inference. During training, we implement this strategy by adjusting $\delta_{train}$ as:
\begin{equation}
    \delta_{train} = \delta_{infer} + (1-\delta_{infer}) \times \exp{(-\frac{\mathrm{update}}{T})},
\end{equation}
where $T$ is a hyperparameter and $\mathrm{update}$ represents the number of updates. During this process, DST gradually adapts to the scenario of generating translations based on prefixes \citep{guo2023glancing}.

\begin{figure*}[t]
\centering
\subfigure[En$\rightarrow $Vi]{
\includegraphics[width=2.07in]{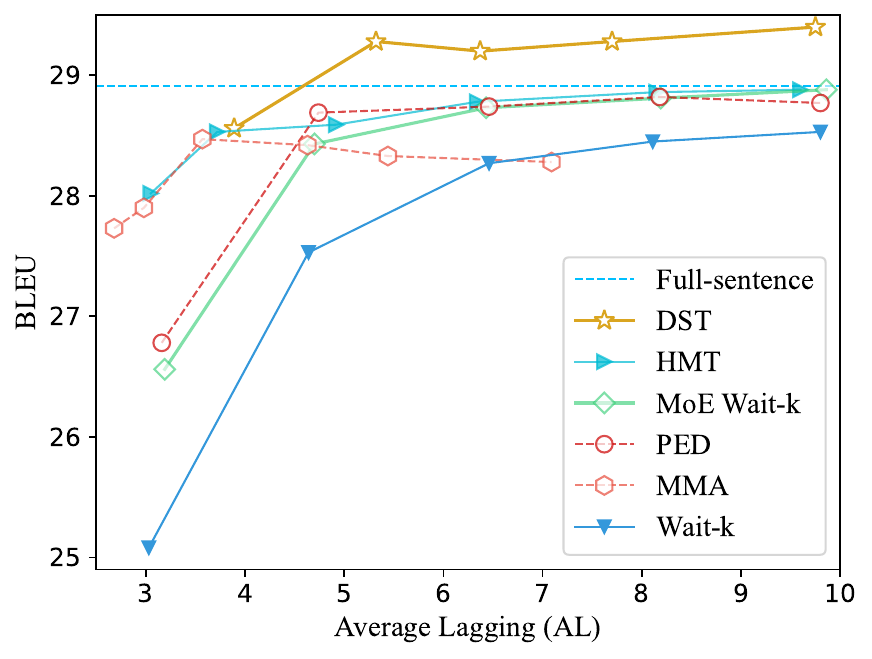}
}\hspace{-0.3cm}
\subfigure[En$\rightarrow $Ro]{
\includegraphics[width=2.04in]{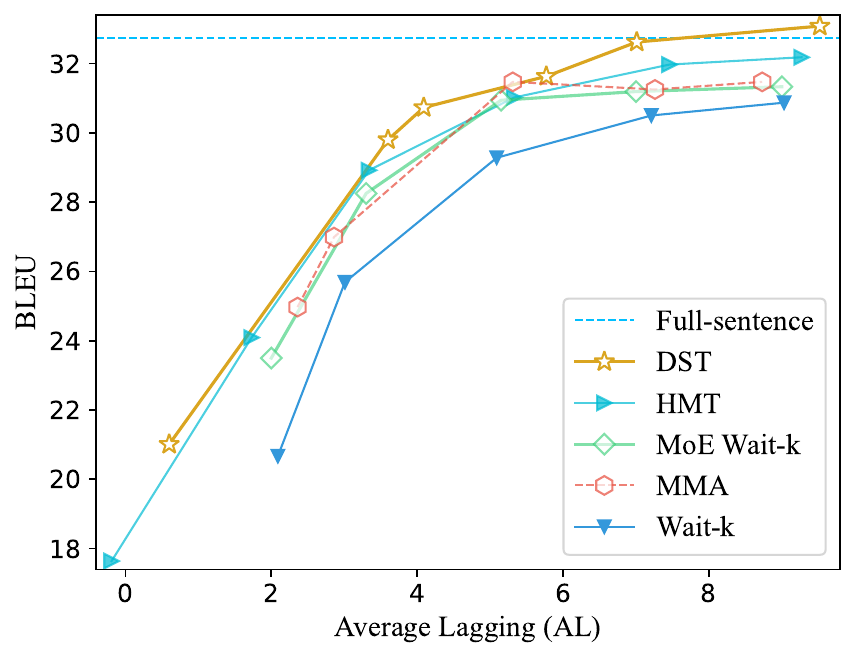}
}\hspace{-0.3cm}
\subfigure[De$\rightarrow $En]{
\includegraphics[width=2.04in]{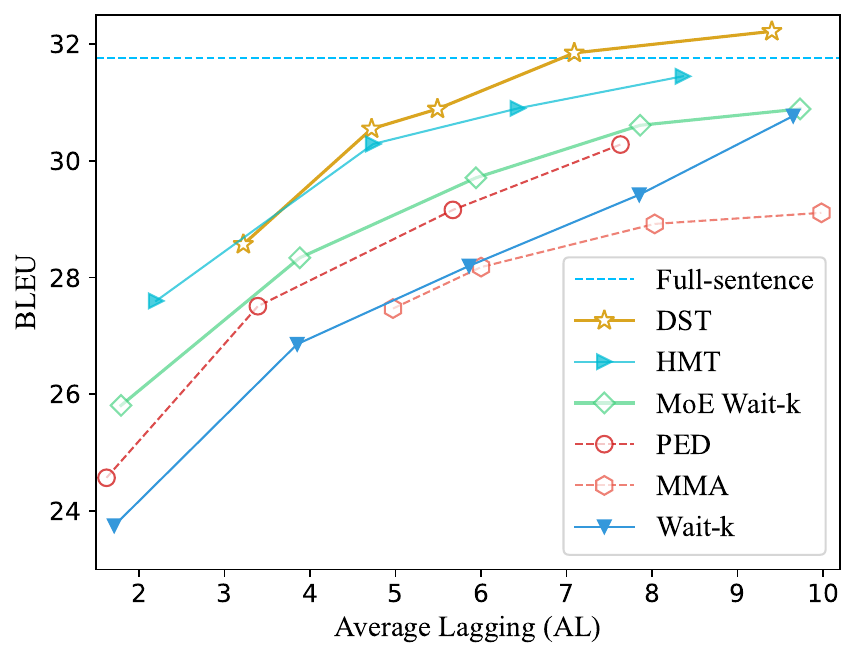}
}

\caption{Comparison of our approach with other SiMT methods on En$\rightarrow$Vi, En$\rightarrow$Ro and De$\rightarrow$En tasks.}
\label{main_res}
\end{figure*}

\section{Experiments}
\subsection{Datasets}
We evaluate our method on three translation tasks.

\textbf{IWSLT15\footnote{\url{https://nlp.stanford.edu/projects/nmt/}} English$\rightarrow$Vietnamese (En$\rightarrow$Vi)} \citep{DBLP:conf/iwslt/CettoloNSBCF15} Consistent with \citet{DBLP:conf/iclr/MaPCPG20}, we replace the tokens occurring less than 5 with $\left \langle unk \right \rangle$. We use TED tst2012 and TED tst2013 as the development set and the test set, respectively.

%\textbf{MuST-C English$\rightarrow$German (En$\rightarrow$De)} We conduct translation on text data \citep{di-gangi-etal-2019-must}. We use dev set for validation and tst-COMMON set for test. We employ a unigram SentencePiece\footnote{\url{https://github.com/google/sentencepiece}} model to build a shared vocabulary with of 10k.

\textbf{WMT16\footnote{\url{www.statmt.org/wmt16/}} English$\rightarrow$Romanian (En$\rightarrow$Ro)} We use newsdev-2016 as validation set and newstest-2016 as test set. The source and target languages employ a shared vocabulary. Other settings are consistent with \citet{DBLP:conf/iclr/Gu0XLS18}.

\textbf{WMT15\footnote{\url{www.statmt.org/wmt15/}} German$\rightarrow$English (De$\rightarrow$En)} We use newstest2013 as validation set and newstest2015 as test set. Following \citet{DBLP:conf/iclr/MaPCPG20}, we apply BPE \citep{sennrich-etal-2016-neural} with 32K subword units and use a shared vocabulary between source and target.

\subsection{System Settings}
Our experiments involve the following methods. Apart from our approach, other methods all employ Encoder-Decoder architecture.

\textbf{Full-sentence} represents the full-sentence translation of Encoder-Decoder architecture.

\textbf{Wait-$k$} \citep{DBLP:conf/acl/MaHXZLZZHLLWW19} reads $k$ tokens and then alternates between writing and reading a token.

%\textbf{Multi-path} \citep{multiPath} trains the Wait-$k$ model by uniformly sampling the latency $k$.

\textbf{MoE Wait-$k$} \citep{DBLP:conf/emnlp/ZhangF21} treats each head as a Wait-$k$ expert and integrates the outputs of all experts to generate translation.

\textbf{MMA} \citep{DBLP:conf/iclr/MaPCPG20} makes each head determine the policy by predicting a Bernoulli variable and generates translation by integrating the results of multiple heads.

\textbf{PED} \citep{DBLP:conf/emnlp/GuoZF22} implements the adaptive policy by integrating post-evaluation into the translation policy.

\textbf{HMT} \citep{DBLP:journals/corr/abs-2303-00257} utilizes the Hidden Markov Model to model the policy as hidden events and translation as observed events, and achieves the current state-of-the-art performance.

\textbf{DST} denotes our method described in Section \ref{all_method}.

All systems are adapted from Fairseq Library \citep{DBLP:conf/naacl/OttEBFGNGA19}. Regarding the compared methods, we apply Transformer-Small (6 layers, 4 heads) for En$\rightarrow$Vi task and Transform-Base (6 layers, 8 heads) for En$\rightarrow$Ro and De$\rightarrow$En tasks. For our approach, we set the number $N$ of layers in DST to 16 and utilize the top 8 layers to determine the translation policy. For the latency constraint, we empirically choose $\epsilon$ as 1. Other detailed system settings are shown in Appendix \ref{system_setting}. We obtain the SiMT models under different latency by adjusting $\delta_{infer}$. To accelerate the convergence and achieve better performance, our approach is fine-tuned from conventional Decoder-only architecture with masked self-attention mechanism \citep{DBLP:conf/nips/VaswaniSPUJGKP17}. We use greedy search during inference and evaluate all methods with latency measured by Average Lagging (AL)  \citep{DBLP:conf/acl/MaHXZLZZHLLWW19} and quality estimated by BLEU \citep{BLEU}.

\subsection{Main Results}
The performance comparison of our approach with other SiMT methods is illustrated in Figure \ref{main_res}, demonstrating that our approach achieves state-of-the-art performance on three tasks. 

\begin{table}[]
\centering
\begin{tabular}{l c c  c} \toprule[1.2pt]
\textbf{$\;\;\;\;\;\;\;\;$Method} & $\delta_{infer}$ & \textbf{AL} & \textbf{BLEU}                                       \\ \cmidrule(lr){1-1} \cmidrule(lr){2-2}\cmidrule(lr){3-4}

DST & 0.3 & \textbf{4.72} &  \textbf{30.55} \\

$\;\;\;\;$w/o $\mathcal{L}_{lat}$ & 0.3 & 7.95 & 31.78 \\

$\;\;\;\;$w/o $\mathcal{L}_{con}$ & 0.3 & 26.5 & 24.05 \\

$\;\;\;\;$w/o CL Strategy& 0.8 & 22.48 & 29.69 \\

 \bottomrule[1pt]
\end{tabular}
\caption{Ablation study on the training method of DST. `w/o $\mathcal{L}_{lat}$' removes the latency constraint. `w/o $\mathcal{L}_{con}$' removes the consistency constraint. `w/o CL Strategy' removes curriculum learning strategy. The experiments are conducted on the De$\rightarrow$En task.}
\label{loss}
\end{table}

Compared to the Wait-$k$ and MoE Wait-$k$ methods, our method brings significant improvement. Both Wait-$k$ and MoE Wait-$k$ employ fixed policy and rely on heuristic rules to determine the source prefixes for translation \citep{DBLP:conf/acl/MaHXZLZZHLLWW19, DBLP:conf/emnlp/ZhangF21}. This usually leads to the omission or redundancy of source information during translation \citep{DualPath}, resulting in degraded performance. In contrast, our method adaptively determines the translation policy by leveraging the correspondence between the source and target prefixes. This adaptability allows our method to achieve superior tradeoffs. Additionally, our method outperforms PED, MMA, and HMT, which belong to the adaptive policy. Previous adaptive methods are based on the Encoder-Decoder architecture, determining the policy based on extracted features from the source and target prefixes \citep{ma-etal-2023-non, DBLP:journals/corr/abs-2303-00257}. Our method relies on the Decoder-only architecture and utilizes the accumulated attention to existing source tokens to assess the sufficiency of translation, resulting in more effective policies. Besides, our method surpasses the performance of Full-sentence translation based on the Encoder-Decoder architecture with lower latency. This can be attributed to the advantages of the decoder-only architecture.

\section{Analysis}
To deepen the understanding of our method, we conduct multiple analyses. The experiments are all performed on the De$\rightarrow$En translations task.
\subsection{Ablation Study}
In order to understand the settings, we conduct ablation studies on the training methods, curriculum learning strategy, and the number of layers. 

Table \ref{loss} presents ablation experiments on the training method, where each setting contributes to the performance of DST. The latency constraint aids the model in acquiring the low-latency policy \citep{ITST}, and the curriculum learning strategy enables the model to adapt to generating translations based on prefixes \citep{guo2023glancing}. The consistency constraint effectively addresses the problems of outlier layers, which ensures consistency across decisions made at different layers \citep{DBLP:conf/iclr/MaPCPG20}. 

\begin{table}[]
\centering
\begin{tabular}{c c c  c} \toprule[1.2pt]
\textbf{$T$} & $\delta_{infer}$ & \textbf{AL} & \textbf{BLEU}                                       \\ \cmidrule(lr){1-1} \cmidrule(lr){2-2}\cmidrule(lr){3-4}

20000 & 0.3 & 2.81 & 25.68 \\

30000 & 0.3 & \textbf{4.72} &  \textbf{30.55} \\

40000 & 0.3 & 4.05 & 29.26 \\

 \bottomrule[1pt]
\end{tabular}
\caption{The performance when DST employs different hyperparameters of curriculum learning strategy. The experiments are performed on the De$\rightarrow$En dataset. }
\label{num_updates}
\end{table}

\begin{table}[]
\centering
\begin{tabular}{c c c  c} \toprule[1.2pt]
\textbf{$N$} & $\delta_{infer}$ & \textbf{AL} & \textbf{BLEU}                                       \\ \cmidrule(lr){1-1} \cmidrule(lr){2-2}\cmidrule(lr){3-4}

12 & 0.3 & 4.23 & 28.99 \\

14 & 0.3 & 4.49 &  29.14 \\

16 & 0.3 & \textbf{4.72} &  \textbf{30.55} \\

 \bottomrule[1pt]
\end{tabular}
\caption{The performance of our method with different number of layers. The results are based on De$\rightarrow$En task.}
\label{layers}
\end{table}

We further explore the impact of hyperparameters in the curriculum learning strategy in Table \ref{num_updates}. A smaller $T$ indicates a quicker transition of the training environment to the scenario of SiMT. We find that transitioning to the scenario of SiMT at an appropriate pace can achieve better tradeoffs between latency and translation quality.

Additionally, we investigate the relationship between the number of layers in DST and its performance in Table \ref{layers}. This suggests that the performance of DST shows incremental improvement as the number of layers increases, highlighting a certain level of scalability.

\begin{table*}[]
\centering
\begin{tabular}{c| c c c c c } \toprule[1.2pt]
\textbf{Method}    & \#Params ($\downarrow$) & Inference Speed-up ($\uparrow$) & Training Speed-up  ($\uparrow$) & AL  & BLEU  \\ \midrule[0.8pt]
Wait-$k$ & 80M & \textbf{1.907} & \textbf{3.982} & 3.85 & 26.86 \\
HMT & 80M & 1.000 & 1.000 & 4.74 & 30.29 \\
DST & \textbf{67M} & 1.391 & 2.029 & \textbf{4.72} & \textbf{30.55} \\
\bottomrule[1pt]
\end{tabular}
\caption{The comparison of model efficiency for different SiMT models. `\#Params' represents the number of trained parameters. `Inference Speed-up' indicates the speed-up ratio of inference compared to the HMT model. `Training Speed-up' represents the speed-up ratio of training in comparison to the HMT model. The bolded data indicates the SiMT models that perform the best in the evaluation metrics. The experiments are conducted on the De$\rightarrow$En task. }
\label{model_efficiency}
\end{table*}

\begin{figure}[t]
    \centering
    \includegraphics[width=2.9in]{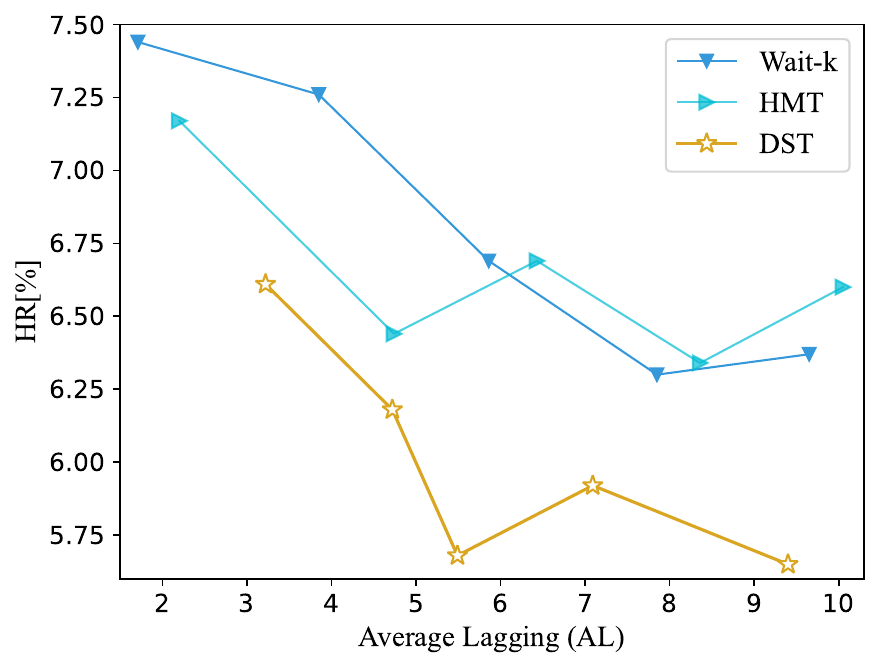}
    \caption{The comparison of hallucination in translations generated by different SiMT models. The results are based on the De$\rightarrow$En dataset.}
    \label{hallucination}
\end{figure}

\subsection{Hallucination in Translation}
To assess the quality of translations generated by different SiMT methods, we evaluate the hallucination in translations. In SiMT, when the model is trained to predict target tokens based on missing essential source information, it is prone to producing hallucinations during inference \citep{guo-etal-2023-learning}. Therefore, the proportion of hallucinations contained in the generated translations also serves as a reflection of the quality of the learned policy during training.

To quantify the hallucinations, we introduce the hallucination rate (HR) \citep{DBLP:conf/emnlp/ChenZKM021}, which measures the percentage of tokens generated that are not related to the source sentence. We then provide its detailed definition. Given the source sentence $\mathbf{x}$, we define the corresponding translation as $\mathbf{{\hat{y}}}$. Subsequently, we can get the alignment set $\mathbf{h}$, which is a set of source-target index pairs ($j$, $i$) where $j^\mathrm{th}$ source token $x_j$ aligns to the $i^\mathrm{th}$ target token $\hat{y}_i$.

The token ${\hat{y}}_i$ in $\mathbf{{\hat{y}}}$ is seen as hallucination ($H(i, \mathbf{h})$=1) if it can not be aligned to any source token:
\begin{equation}
    H(i, \mathbf{h})\!=\! \mathbbm{1} [ \{ (j, i) \in \mathbf{h} \} = \varnothing ].
\end{equation}
Therefore, the hallucination rate can be defined as:
\begin{equation}
    HR(\mathbf{x}, \mathbf{{\hat{y}}}, \mathbf{h})\!=\!
    \frac{1}
    {|\mathbf{{\hat{y}}}|}
    \sum\limits_{i=1}^{|\mathbf{{\hat{y}}}|}
    H(i, \mathbf{h}).
\end{equation}

As illustrated in Figure \ref{hallucination}, our model demonstrates a lower likelihood of generating hallucinations at all latency, indicating increased reliability in generated translations. Compared to the Wait-$k$ approach, our method allows for flexible utilization of essential source information by adjusting its policy and obtains a lower hallucination rate. Moreover, DST considers a more extensive range of translation policies than HMT during training \citep{DBLP:journals/corr/abs-2303-00257}. This provides possibilities for DST to acquire more effective policies, thereby achieving better tradeoffs between latency and translation quality.

\subsection{Model Efficiency}
In addition to evaluating the hallucinations in translation, we also investigate the model efficiency of different SiMT methods. As shown in Table \ref{model_efficiency}, our method obtains superior performance with relatively high efficiency.

Compared to the SiMT models based on the Encoder-Decoder architecture, our model, utilizing a Decoder-only architecture, achieves superior performance with a relatively smaller number of parameters. Combining the results from Table \ref{layers}, it reflects the parameter efficiency and scalability of the Decoder-only architecture \citep{fu2023decoderonly}. 

Besides, we evaluate the training and inference efficiency of different SiMT models. All related experiments are conducted on NVIDIA GeForce RTX 3090. In terms of training speed, our approach is lower than Wait-$k$ policy but higher than the adaptive policy HMT. Compared to Wait-$k$ policy, our method involves more computations about attention, therefore requiring more training costs. Furthermore, HMT significantly increases the training complexity by expanding the target sentences several times during training. During inference, while our method is slightly slower than Wait-$k$, the inference speed of our method can still reach 78 tokens per second, which can fully meet the application requirements. 

\begin{table}[]
\centering
\begin{tabular}{p{2cm}<{\centering} p{2cm}<{\centering} p{2cm}<{\centering}} 
\toprule[1.5pt]
    Method     & CAAL & SacreBLEU   \\
    \midrule

    \multirow{2}{*}{Wait-$k$}    &1235.72     &25.83 \\
                &1704.95     &27.30 \\
    \midrule
    \multirow{2}{*}{DST}     &1672.48     &27.53  \\
    &1702.00     &28.17  \\
\bottomrule[1pt]
\end{tabular}
\caption{Comparison of different methods on MuST-C English$\rightarrow$German task. This latency metric CAAL is measured in millisecond.}
\label{ende}
\end{table}

%To further evaluate the applicability of our approach, we consider both the machine inference time and the delay caused by waiting for the source information. We evaluate our method and Wait-$k$ policy, and detailed experimental settings can be found in Appendix \ref{computation_aware}.
In addition to comparing the training and inference efficiency of SiMT methods in Table \ref{model_efficiency}, we also consider model inference time and delay caused by waiting for source information concurrently. To evaluate the performance of the SiMT models in scenarios that closely resemble real-world conditions, we compute Computation-Aware Average Lagging (CAAL) \citep{ma-etal-2020-simulmt} for different methods utilizing MuST-C English$\rightarrow$German dataset \citep{di-gangi-etal-2019-must}. Unlike \citet{ma-etal-2020-simulmt}, our approach does not take direct speech input but rather uses the text corresponding to the speech. Therefore, we use \texttt{whisper-large-v3}\footnote{\url{https://huggingface.co/openai/whisper-large-v3}} to align the text with the speech, considering both the length of the corresponding speech sequence and the actual machine inference time when calculating CAAL. Our settings use the Transformer-Small architecture and evaluate the performance of SiMT methods using CAAL and SacreBLEU \citep{post-2018-call}. The results are shown in Table \ref{ende}. Although the inference speed of our method is slightly slower than the Wait-k policy in Table \ref{model_efficiency}, it still achieves better performance in more realistic scenarios.

\begin{table}[]
\centering
\begin{tabular}{c c c  c} \toprule[1.2pt]
\textbf{Method} & $\delta_{infer}$ & \textbf{AL} & \textbf{BLEU}                                       \\ \cmidrule(lr){1-1} \cmidrule(lr){2-2}\cmidrule(lr){3-4}

Expected-Allocation & 0.3 & \textbf{4.72} &  \textbf{30.55} \\

Max-Allocation & 0.3 & 13.73 & 13.79 \\

 \bottomrule[1pt]
\end{tabular}
\caption{The performance of the SiMT model when using different strategies to allocate attention to the source prefixes in the Streaming Self-Attention (SSA) mechanism. `Expected-Allocation' represents that SSA allocates attention to all possible source prefixes. `Max-Allocation' signifies the allocation of source attention probability to the most probable source prefix. }
\label{soft-hard}
\end{table}

Therefore, our Decoder-only method can achieve state-of-the-art performance with higher parameter utilization, and relatively faster inference and training speed.

\subsection{Analysis of Streaming Self-Attention}

After evaluating the overall performance of our method, we delve into assessing the effectiveness of the proposed Streaming Self-Attention (SSA) mechanism. SSA initially allocates attention probability to all possible source prefixes and subsequently combines the allocated attention with the soft-attention to determine the expected attention for source tokens, referred to as Expected-Allocation. To demonstrate the effectiveness of SSA, we also explore a Max-Allocation strategy, where the attention probability is allocated solely to the most probable source prefix.

As illustrated in Table \ref{soft-hard}, the Expected-Allocation method yields better latency-quality tradeoffs. The Expected-Allocation method integrates the translation policy with translation capability. It aids the model in learning the correlation between source and target prefixes, resulting in enhanced translation policies and overall performance. This underscores the importance of attention probability allocation in SSA.

\section{Related Work}
Simultaneous machine translation (SiMT) begins generating translation before reading the entire source sentence. To ensure the quality of generated translations and meet the latency requirements, it is necessary to determine the suitable source prefixes for translation. The process of determining the source prefixes for target tokens is referred to as the policy \citep{zhang-2024}. Depending on whether the model decides the policy utilizing the correspondence between source and target prefixes, SiMT methods can be broadly categorized into fixed policy and adaptive policy.

For fixed policy, the SiMT model reads or generates tokens according to the heuristic rules. \citet{DBLP:conf/acl/MaHXZLZZHLLWW19} proposes Wait-$k$ policy, which reads $k$ source tokens first, and then writes and reads one token alternately. \citet{multiPath} introduces the unidirectional encoder, where the preceding source tokens cannot attend to subsequent words. \citet{DBLP:conf/emnlp/ZhangF21} proposes MoE Wait-$k$, which assigns a Wait-$k$ policy to each head and combines the results from multiple heads to generate translations. There are some methods that enhance the SiMT capability by modifying the reference \citep{chen-etal-2021-improving-simultaneous, guo-etal-2023-simultaneous}. However, the fixed policy requires the SiMT model to generate tokens even when source information is insufficient, leading to a decline in performance of SiMT.

%However, traditional bidirectional encoder \citep{DBLP:conf/nips/VaswaniSPUJGKP17} requires the model to re-encode previously read source tokens each time a new source token is input. 

Unlike fixed policy, adaptive policy can dynamically decide whether to read or generate tokens based on the correspondence between source prefix and target prefix. \citet{DBLP:conf/acl/ZhengLZMLH20} implements the adaptive policy through a composition of several fixed policies. \citet{DBLP:conf/emnlp/MiaoBS21} proposes a generative framework, which uses a re-parameterized Poisson prior to regularise the policy. \citet{ITST} models the contribution of each source token to the target tokens through optimal transport and determines the policy by accumulating the contributions of input source tokens. \citet{DBLP:journals/corr/abs-2303-00257} utilizes the Hidden Markov Model to model the translation policy as the hidden events and the translation as the observed events. \citet{ma-etal-2023-non, ma-etal-2024-nonautoregressive} first proposes to conduct SiMT with non-autoregressive architecture and achieves good results. However, previous SiMT methods all employ the Encoder-Decoder architecture.
%\citet{DBLP:conf/iclr/MaPCPG20} allows each head to determine its policy by predicting a Bernoulli variable, and integrates the decision results from multiple heads.

With the rise of language models, the Decoder-only architecture has demonstrated superior performance across various tasks \citep{touvron2023llama, gemmateam2024gemma, jiang2024mixtral}. The Decoder-only architecture has the advantages of good scalability and parameter efficiency. More importantly, its unidirectional encoding nature aligns greatly with streaming input. Therefore, we explore the application of the Decoder-only architecture in SiMT and propose the first Decoder-only SiMT model.

\section{Conclusion}
In this paper, we propose the first Decoder-only SiMT model, which leverages the characteristics of the Decoder-only architecture to implement the adaptive policy. Experiments show that our method achieves state-of-the-art performance and exhibits excellent scalability and model efficiency.

\section*{Limitations}
Regarding the system settings, we investigate the impact of the number of layers and training methods on the performance of our DST method due to resource constraints. We think that further exploration of system settings could potentially yield even better results. We leave the exploration for future work.

\section*{Acknowledgements}

This paper is supported by National Natural Science Foundation of China (Grant No. 62376260). We thank all the anonymous reviewers for their valuable feedback and thorough reviews.

\bibliography{custom}

\begin{thebibliography}{42}
\expandafter\ifx\csname natexlab\endcsname\relax\def\natexlab#1{#1}\fi

\bibitem[{Bengio et~al.(2009)Bengio, Louradour, Collobert, and Weston}]{bengio2009curriculum}
Yoshua Bengio, J{\'e}r{\^o}me Louradour, Ronan Collobert, and Jason Weston. 2009.
\newblock Curriculum learning.
\newblock In \emph{Proceedings of the 26th annual international conference on machine learning}, pages 41--48.

\bibitem[{Cettolo et~al.(2015)Cettolo, Niehues, St{\"{u}}ker, Bentivogli, Cattoni, and Federico}]{DBLP:conf/iwslt/CettoloNSBCF15}
Mauro Cettolo, Jan Niehues, Sebastian St{\"{u}}ker, Luisa Bentivogli, Roldano Cattoni, and Marcello Federico. 2015.
\newblock \href {https://aclanthology.org/2015.iwslt-evaluation.1} {The {IWSLT} 2015 evaluation campaign}.
\newblock In \emph{Proceedings of the 12th International Workshop on Spoken Language Translation: Evaluation Campaign@IWSLT 2015, Da Nang, Vietnam, December 3-4, 2015}.

\bibitem[{Chen et~al.(2021{\natexlab{a}})Chen, Zheng, Kita, Ma, and Huang}]{DBLP:conf/emnlp/ChenZKM021}
Junkun Chen, Renjie Zheng, Atsuhito Kita, Mingbo Ma, and Liang Huang. 2021{\natexlab{a}}.
\newblock \href {https://doi.org/10.18653/v1/2021.emnlp-main.473} {Improving simultaneous translation by incorporating pseudo-references with fewer reorderings}.
\newblock In \emph{Proceedings of the 2021 Conference on Empirical Methods in Natural Language Processing, {EMNLP} 2021, Virtual Event / Punta Cana, Dominican Republic, 7-11 November, 2021}, pages 5857--5864. Association for Computational Linguistics.

\bibitem[{Chen et~al.(2021{\natexlab{b}})Chen, Zheng, Kita, Ma, and Huang}]{chen-etal-2021-improving-simultaneous}
Junkun Chen, Renjie Zheng, Atsuhito Kita, Mingbo Ma, and Liang Huang. 2021{\natexlab{b}}.
\newblock \href {https://doi.org/10.18653/v1/2021.emnlp-main.473} {Improving simultaneous translation by incorporating pseudo-references with fewer reorderings}.
\newblock In \emph{Proceedings of the 2021 Conference on Empirical Methods in Natural Language Processing}, pages 5857--5864, Online and Punta Cana, Dominican Republic. Association for Computational Linguistics.

\bibitem[{Dalvi et~al.(2018)Dalvi, Durrani, Sajjad, and Vogel}]{dalvi-etal-2018-incremental}
Fahim Dalvi, Nadir Durrani, Hassan Sajjad, and Stephan Vogel. 2018.
\newblock \href {https://doi.org/10.18653/v1/N18-2079} {Incremental decoding and training methods for simultaneous translation in neural machine translation}.
\newblock In \emph{Proceedings of the 2018 Conference of the North {A}merican Chapter of the Association for Computational Linguistics: Human Language Technologies, Volume 2 (Short Papers)}, pages 493--499, New Orleans, Louisiana. Association for Computational Linguistics.

\bibitem[{Di~Gangi et~al.(2019)Di~Gangi, Cattoni, Bentivogli, Negri, and Turchi}]{di-gangi-etal-2019-must}
Mattia~A. Di~Gangi, Roldano Cattoni, Luisa Bentivogli, Matteo Negri, and Marco Turchi. 2019.
\newblock \href {https://doi.org/10.18653/v1/N19-1202} {{M}u{ST}-{C}: a {M}ultilingual {S}peech {T}ranslation {C}orpus}.
\newblock In \emph{Proceedings of the 2019 Conference of the North {A}merican Chapter of the Association for Computational Linguistics: Human Language Technologies, Volume 1 (Long and Short Papers)}, pages 2012--2017, Minneapolis, Minnesota. Association for Computational Linguistics.

\bibitem[{Elbayad et~al.(2020)Elbayad, Besacier, and Verbeek}]{multiPath}
Maha Elbayad, Laurent Besacier, and Jakob Verbeek. 2020.
\newblock \href {https://doi.org/10.21437/Interspeech.2020-1241} {Efficient wait-k models for simultaneous machine translation}.
\newblock In \emph{Interspeech 2020, 21st Annual Conference of the International Speech Communication Association, Virtual Event, Shanghai, China, 25-29 October 2020}, pages 1461--1465. {ISCA}.

\bibitem[{Fu et~al.(2023)Fu, Lam, Yu, So, Hu, Liu, and Collier}]{fu2023decoderonly}
Zihao Fu, Wai Lam, Qian Yu, Anthony Man-Cho So, Shengding Hu, Zhiyuan Liu, and Nigel Collier. 2023.
\newblock \href {http://arxiv.org/abs/2304.04052} {Decoder-only or encoder-decoder? interpreting language model as a regularized encoder-decoder}.

\bibitem[{Gu et~al.(2018)Gu, Bradbury, Xiong, Li, and Socher}]{DBLP:conf/iclr/Gu0XLS18}
Jiatao Gu, James Bradbury, Caiming Xiong, Victor O.~K. Li, and Richard Socher. 2018.
\newblock \href {https://openreview.net/forum?id=B1l8BtlCb} {Non-autoregressive neural machine translation}.
\newblock In \emph{6th International Conference on Learning Representations, {ICLR} 2018, Vancouver, BC, Canada, April 30 - May 3, 2018, Conference Track Proceedings}. OpenReview.net.

\bibitem[{Gu et~al.(2017)Gu, Neubig, Cho, and Li}]{reinforcement}
Jiatao Gu, Graham Neubig, Kyunghyun Cho, and Victor~O.K. Li. 2017.
\newblock \href {https://aclanthology.org/E17-1099} {Learning to translate in real-time with neural machine translation}.
\newblock In \emph{Proceedings of the 15th Conference of the {E}uropean Chapter of the Association for Computational Linguistics: Volume 1, Long Papers}, pages 1053--1062, Valencia, Spain. Association for Computational Linguistics.

\bibitem[{Guo et~al.(2022)Guo, Zhang, and Feng}]{DBLP:conf/emnlp/GuoZF22}
Shoutao Guo, Shaolei Zhang, and Yang Feng. 2022.
\newblock \href {https://aclanthology.org/2022.findings-emnlp.167} {Turning fixed to adaptive: Integrating post-evaluation into simultaneous machine translation}.
\newblock In \emph{Findings of the Association for Computational Linguistics: {EMNLP} 2022, Abu Dhabi, United Arab Emirates, December 7-11, 2022}, pages 2264--2278. Association for Computational Linguistics.

\bibitem[{Guo et~al.(2023{\natexlab{a}})Guo, Zhang, and Feng}]{guo-etal-2023-learning}
Shoutao Guo, Shaolei Zhang, and Yang Feng. 2023{\natexlab{a}}.
\newblock \href {https://doi.org/10.18653/v1/2023.acl-long.130} {Learning optimal policy for simultaneous machine translation via binary search}.
\newblock In \emph{Proceedings of the 61st Annual Meeting of the Association for Computational Linguistics (Volume 1: Long Papers)}, pages 2318--2333, Toronto, Canada. Association for Computational Linguistics.

\bibitem[{Guo et~al.(2023{\natexlab{b}})Guo, Zhang, and Feng}]{guo-etal-2023-simultaneous}
Shoutao Guo, Shaolei Zhang, and Yang Feng. 2023{\natexlab{b}}.
\newblock \href {https://doi.org/10.18653/v1/2023.findings-emnlp.202} {Simultaneous machine translation with tailored reference}.
\newblock In \emph{Findings of the Association for Computational Linguistics: EMNLP 2023}, pages 3070--3084, Singapore. Association for Computational Linguistics.

\bibitem[{Guo et~al.(2024)Guo, Zhang, and Feng}]{guo2023glancing}
Shoutao Guo, Shaolei Zhang, and Yang Feng. 2024.
\newblock \href {https://doi.org/10.1109/ICASSP48485.2024.10446517} {Glancing future for simultaneous machine translation}.
\newblock In \emph{ICASSP 2024 - 2024 IEEE International Conference on Acoustics, Speech and Signal Processing (ICASSP)}, pages 11386--11390.

\bibitem[{Jiang et~al.(2024)Jiang, Sablayrolles, Roux, Mensch, Savary, Bamford, Chaplot, de~las Casas, Hanna, Bressand, Lengyel, Bour, Lample, Lavaud, Saulnier, Lachaux, Stock, Subramanian, Yang, Antoniak, Scao, Gervet, Lavril, Wang, Lacroix, and Sayed}]{jiang2024mixtral}
Albert~Q. Jiang, Alexandre Sablayrolles, Antoine Roux, Arthur Mensch, Blanche Savary, Chris Bamford, Devendra~Singh Chaplot, Diego de~las Casas, Emma~Bou Hanna, Florian Bressand, Gianna Lengyel, Guillaume Bour, Guillaume Lample, Lélio~Renard Lavaud, Lucile Saulnier, Marie-Anne Lachaux, Pierre Stock, Sandeep Subramanian, Sophia Yang, Szymon Antoniak, Teven~Le Scao, Théophile Gervet, Thibaut Lavril, Thomas Wang, Timothée Lacroix, and William~El Sayed. 2024.
\newblock \href {http://arxiv.org/abs/2401.04088} {Mixtral of experts}.

\bibitem[{Liu et~al.(2023)Liu, Gao, Chen, Zhao, and Wen}]{liu-etal-2023-enhancing-scalability}
Peiyu Liu, Ze-Feng Gao, Yushuo Chen, Xin Zhao, and Ji-Rong Wen. 2023.
\newblock \href {https://doi.org/10.18653/v1/2023.findings-emnlp.920} {Enhancing scalability of pre-trained language models via efficient parameter sharing}.
\newblock In \emph{Findings of the Association for Computational Linguistics: EMNLP 2023}, pages 13771--13785, Singapore. Association for Computational Linguistics.

\bibitem[{Ma et~al.(2019)Ma, Huang, Xiong, Zheng, Liu, Zheng, Zhang, He, Liu, Li, Wu, and Wang}]{DBLP:conf/acl/MaHXZLZZHLLWW19}
Mingbo Ma, Liang Huang, Hao Xiong, Renjie Zheng, Kaibo Liu, Baigong Zheng, Chuanqiang Zhang, Zhongjun He, Hairong Liu, Xing Li, Hua Wu, and Haifeng Wang. 2019.
\newblock \href {https://doi.org/10.18653/v1/p19-1289} {{STACL:} simultaneous translation with implicit anticipation and controllable latency using prefix-to-prefix framework}.
\newblock In \emph{Proceedings of the 57th Conference of the Association for Computational Linguistics, {ACL} 2019, Florence, Italy, July 28- August 2, 2019, Volume 1: Long Papers}, pages 3025--3036. Association for Computational Linguistics.

\bibitem[{Ma et~al.(2020{\natexlab{a}})Ma, Pino, and Koehn}]{ma-etal-2020-simulmt}
Xutai Ma, Juan Pino, and Philipp Koehn. 2020{\natexlab{a}}.
\newblock \href {https://aclanthology.org/2020.aacl-main.58} {{S}imul{MT} to {S}imul{ST}: Adapting simultaneous text translation to end-to-end simultaneous speech translation}.
\newblock In \emph{Proceedings of the 1st Conference of the Asia-Pacific Chapter of the Association for Computational Linguistics and the 10th International Joint Conference on Natural Language Processing}, pages 582--587, Suzhou, China. Association for Computational Linguistics.

\bibitem[{Ma et~al.(2020{\natexlab{b}})Ma, Pino, Cross, Puzon, and Gu}]{DBLP:conf/iclr/MaPCPG20}
Xutai Ma, Juan~Miguel Pino, James Cross, Liezl Puzon, and Jiatao Gu. 2020{\natexlab{b}}.
\newblock \href {https://openreview.net/forum?id=Hyg96gBKPS} {Monotonic multihead attention}.
\newblock In \emph{8th International Conference on Learning Representations, {ICLR} 2020, Addis Ababa, Ethiopia, April 26-30, 2020}. OpenReview.net.

\bibitem[{Ma et~al.(2024)Ma, Fang, Zhang, Guo, Feng, and Zhang}]{ma-etal-2024-nonautoregressive}
Zhengrui Ma, Qingkai Fang, Shaolei Zhang, Shoutao Guo, Yang Feng, and Min Zhang. 2024.
\newblock A non-autoregressive generation framework for end-to-end simultaneous speech-to-any translation.
\newblock In \emph{Proceedings of the 62th Annual Meeting of the Association for Computational Linguistics (Long Papers)}, Bangkok, Thailand. Association for Computational Linguistics.

\bibitem[{Ma et~al.(2023)Ma, Zhang, Guo, Shao, Zhang, and Feng}]{ma-etal-2023-non}
Zhengrui Ma, Shaolei Zhang, Shoutao Guo, Chenze Shao, Min Zhang, and Yang Feng. 2023.
\newblock \href {https://doi.org/10.18653/v1/2023.emnlp-main.314} {Non-autoregressive streaming transformer for simultaneous translation}.
\newblock In \emph{Proceedings of the 2023 Conference on Empirical Methods in Natural Language Processing}, pages 5177--5190, Singapore. Association for Computational Linguistics.

\bibitem[{Miao et~al.(2021)Miao, Blunsom, and Specia}]{DBLP:conf/emnlp/MiaoBS21}
Yishu Miao, Phil Blunsom, and Lucia Specia. 2021.
\newblock \href {https://doi.org/10.18653/v1/2021.emnlp-main.536} {A generative framework for simultaneous machine translation}.
\newblock In \emph{Proceedings of the 2021 Conference on Empirical Methods in Natural Language Processing, {EMNLP} 2021, Virtual Event / Punta Cana, Dominican Republic, 7-11 November, 2021}, pages 6697--6706. Association for Computational Linguistics.

\bibitem[{Ott et~al.(2019)Ott, Edunov, Baevski, Fan, Gross, Ng, Grangier, and Auli}]{DBLP:conf/naacl/OttEBFGNGA19}
Myle Ott, Sergey Edunov, Alexei Baevski, Angela Fan, Sam Gross, Nathan Ng, David Grangier, and Michael Auli. 2019.
\newblock \href {https://doi.org/10.18653/v1/n19-4009} {fairseq: {A} fast, extensible toolkit for sequence modeling}.
\newblock In \emph{Proceedings of the 2019 Conference of the North American Chapter of the Association for Computational Linguistics: Human Language Technologies, {NAACL-HLT} 2019, Minneapolis, MN, USA, June 2-7, 2019, Demonstrations}, pages 48--53. Association for Computational Linguistics.

\bibitem[{Papineni et~al.(2002)Papineni, Roukos, Ward, and Zhu}]{BLEU}
Kishore Papineni, Salim Roukos, Todd Ward, and Wei-Jing Zhu. 2002.
\newblock \href {https://doi.org/10.3115/1073083.1073135} {{B}leu: a method for automatic evaluation of machine translation}.
\newblock In \emph{Proceedings of the 40th Annual Meeting of the Association for Computational Linguistics}, pages 311--318, Philadelphia, Pennsylvania, USA. Association for Computational Linguistics.

\bibitem[{Post(2018)}]{post-2018-call}
Matt Post. 2018.
\newblock \href {https://doi.org/10.18653/v1/W18-6319} {A call for clarity in reporting {BLEU} scores}.
\newblock In \emph{Proceedings of the Third Conference on Machine Translation: Research Papers}, pages 186--191, Brussels, Belgium. Association for Computational Linguistics.

\bibitem[{Sennrich et~al.(2016)Sennrich, Haddow, and Birch}]{sennrich-etal-2016-neural}
Rico Sennrich, Barry Haddow, and Alexandra Birch. 2016.
\newblock \href {https://doi.org/10.18653/v1/P16-1162} {Neural machine translation of rare words with subword units}.
\newblock In \emph{Proceedings of the 54th Annual Meeting of the Association for Computational Linguistics (Volume 1: Long Papers)}, pages 1715--1725, Berlin, Germany. Association for Computational Linguistics.

\bibitem[{Team et~al.(2024)Team, Mesnard, Hardin, Dadashi, Bhupatiraju, Pathak, Sifre, Rivière, Kale, Love, Tafti, Hussenot, Sessa, Chowdhery, Roberts, Barua, Botev, Castro-Ros, Slone, Héliou, Tacchetti, Bulanova, Paterson, Tsai, Shahriari, Lan, Choquette-Choo, Crepy, Cer, Ippolito, Reid, Buchatskaya, Ni, Noland, Yan, Tucker, Muraru, Rozhdestvenskiy, Michalewski, Tenney, Grishchenko, Austin, Keeling, Labanowski, Lespiau, Stanway, Brennan, Chen, Ferret, Chiu, Mao-Jones, Lee, Yu, Millican, Sjoesund, Lee, Dixon, Reid, Mikuła, Wirth, Sharman, Chinaev, Thain, Bachem, Chang, Wahltinez, Bailey, Michel, Yotov, Chaabouni, Comanescu, Jana, Anil, McIlroy, Liu, Mullins, Smith, Borgeaud, Girgin, Douglas, Pandya, Shakeri, De, Klimenko, Hennigan, Feinberg, Stokowiec, hui Chen, Ahmed, Gong, Warkentin, Peran, Giang, Farabet, Vinyals, Dean, Kavukcuoglu, Hassabis, Ghahramani, Eck, Barral, Pereira, Collins, Joulin, Fiedel, Senter, Andreev, and Kenealy}]{gemmateam2024gemma}
Gemma Team, Thomas Mesnard, Cassidy Hardin, Robert Dadashi, Surya Bhupatiraju, Shreya Pathak, Laurent Sifre, Morgane Rivière, Mihir~Sanjay Kale, Juliette Love, Pouya Tafti, Léonard Hussenot, Pier~Giuseppe Sessa, Aakanksha Chowdhery, Adam Roberts, Aditya Barua, Alex Botev, Alex Castro-Ros, Ambrose Slone, Amélie Héliou, Andrea Tacchetti, Anna Bulanova, Antonia Paterson, Beth Tsai, Bobak Shahriari, Charline~Le Lan, Christopher~A. Choquette-Choo, Clément Crepy, Daniel Cer, Daphne Ippolito, David Reid, Elena Buchatskaya, Eric Ni, Eric Noland, Geng Yan, George Tucker, George-Christian Muraru, Grigory Rozhdestvenskiy, Henryk Michalewski, Ian Tenney, Ivan Grishchenko, Jacob Austin, James Keeling, Jane Labanowski, Jean-Baptiste Lespiau, Jeff Stanway, Jenny Brennan, Jeremy Chen, Johan Ferret, Justin Chiu, Justin Mao-Jones, Katherine Lee, Kathy Yu, Katie Millican, Lars~Lowe Sjoesund, Lisa Lee, Lucas Dixon, Machel Reid, Maciej Mikuła, Mateo Wirth, Michael Sharman, Nikolai Chinaev, Nithum Thain, Olivier Bachem,
  Oscar Chang, Oscar Wahltinez, Paige Bailey, Paul Michel, Petko Yotov, Rahma Chaabouni, Ramona Comanescu, Reena Jana, Rohan Anil, Ross McIlroy, Ruibo Liu, Ryan Mullins, Samuel~L Smith, Sebastian Borgeaud, Sertan Girgin, Sholto Douglas, Shree Pandya, Siamak Shakeri, Soham De, Ted Klimenko, Tom Hennigan, Vlad Feinberg, Wojciech Stokowiec, Yu~hui Chen, Zafarali Ahmed, Zhitao Gong, Tris Warkentin, Ludovic Peran, Minh Giang, Clément Farabet, Oriol Vinyals, Jeff Dean, Koray Kavukcuoglu, Demis Hassabis, Zoubin Ghahramani, Douglas Eck, Joelle Barral, Fernando Pereira, Eli Collins, Armand Joulin, Noah Fiedel, Evan Senter, Alek Andreev, and Kathleen Kenealy. 2024.
\newblock \href {http://arxiv.org/abs/2403.08295} {Gemma: Open models based on gemini research and technology}.

\bibitem[{Touvron et~al.(2023)Touvron, Martin, Stone, Albert, Almahairi, Babaei, Bashlykov, Batra, Bhargava, Bhosale et~al.}]{touvron2023llama}
Hugo Touvron, Louis Martin, Kevin Stone, Peter Albert, Amjad Almahairi, Yasmine Babaei, Nikolay Bashlykov, Soumya Batra, Prajjwal Bhargava, Shruti Bhosale, et~al. 2023.
\newblock Llama 2: Open foundation and fine-tuned chat models.
\newblock \emph{arXiv preprint arXiv:2307.09288}.

\bibitem[{Vaswani et~al.(2017)Vaswani, Shazeer, Parmar, Uszkoreit, Jones, Gomez, Kaiser, and Polosukhin}]{DBLP:conf/nips/VaswaniSPUJGKP17}
Ashish Vaswani, Noam Shazeer, Niki Parmar, Jakob Uszkoreit, Llion Jones, Aidan~N. Gomez, Lukasz Kaiser, and Illia Polosukhin. 2017.
\newblock \href {https://proceedings.neurips.cc/paper/2017/hash/3f5ee243547dee91fbd053c1c4a845aa-Abstract.html} {Attention is all you need}.
\newblock In \emph{Advances in Neural Information Processing Systems 30: Annual Conference on Neural Information Processing Systems 2017, December 4-9, 2017, Long Beach, CA, {USA}}, pages 5998--6008.

\bibitem[{Wang et~al.(2024)Wang, Vu, Shareghi, and Haffari}]{wang2024conversational}
Minghan Wang, Thuy-Trang Vu, Ehsan Shareghi, and Gholamreza Haffari. 2024.
\newblock \href {http://arxiv.org/abs/2402.10552} {Conversational simulmt: Efficient simultaneous translation with large language models}.

\bibitem[{Yang et~al.(2020)Yang, Wang, Shi, Tadepalli, Lee, and Tu}]{yang-etal-2020-sub}
Yilin Yang, Longyue Wang, Shuming Shi, Prasad Tadepalli, Stefan Lee, and Zhaopeng Tu. 2020.
\newblock \href {https://doi.org/10.18653/v1/2020.findings-emnlp.432} {On the sub-layer functionalities of transformer decoder}.
\newblock In \emph{Findings of the Association for Computational Linguistics: EMNLP 2020}, pages 4799--4811, Online. Association for Computational Linguistics.

\bibitem[{Zhang et~al.(2024)Zhang, Fang, Guo, Ma, Zhang, and Feng}]{zhang-2024}
Shaolei Zhang, QingKai Fang, Shoutao Guo, Zhengrui Ma, Min Zhang, and Yang Feng. 2024.
\newblock \href {https://arxiv.org/abs/2406.03049} {Streamspeech: Simultaneous speech-to-speech translation with multi-task learning}.
\newblock In \emph{Proceedings of the 62th Annual Meeting of the Association for Computational Linguistics (Long Papers)}, Bangkok, Thailand. Association for Computational Linguistics.

\bibitem[{Zhang and Feng(2021)}]{DBLP:conf/emnlp/ZhangF21}
Shaolei Zhang and Yang Feng. 2021.
\newblock \href {https://doi.org/10.18653/v1/2021.emnlp-main.581} {Universal simultaneous machine translation with mixture-of-experts wait-k policy}.
\newblock In \emph{Proceedings of the 2021 Conference on Empirical Methods in Natural Language Processing, {EMNLP} 2021, Virtual Event / Punta Cana, Dominican Republic, 7-11 November, 2021}, pages 7306--7317. Association for Computational Linguistics.

\bibitem[{Zhang and Feng(2022{\natexlab{a}})}]{gaussian}
Shaolei Zhang and Yang Feng. 2022{\natexlab{a}}.
\newblock \href {https://doi.org/10.18653/v1/2022.findings-acl.238} {{G}aussian multi-head attention for simultaneous machine translation}.
\newblock In \emph{Findings of the Association for Computational Linguistics: ACL 2022}, pages 3019--3030, Dublin, Ireland. Association for Computational Linguistics.

\bibitem[{Zhang and Feng(2022{\natexlab{b}})}]{ITST}
Shaolei Zhang and Yang Feng. 2022{\natexlab{b}}.
\newblock \href {https://aclanthology.org/2022.emnlp-main.65} {Information-transport-based policy for simultaneous translation}.
\newblock In \emph{Proceedings of the 2022 Conference on Empirical Methods in Natural Language Processing}, pages 992--1013, Abu Dhabi, United Arab Emirates. Association for Computational Linguistics.

\bibitem[{Zhang and Feng(2022{\natexlab{c}})}]{DualPath}
Shaolei Zhang and Yang Feng. 2022{\natexlab{c}}.
\newblock \href {https://doi.org/10.18653/v1/2022.acl-long.176} {Modeling dual read/write paths for simultaneous machine translation}.
\newblock In \emph{Proceedings of the 60th Annual Meeting of the Association for Computational Linguistics (Volume 1: Long Papers)}, pages 2461--2477, Dublin, Ireland. Association for Computational Linguistics.

\bibitem[{Zhang and Feng(2022{\natexlab{d}})}]{zhang-feng-2022-reducing}
Shaolei Zhang and Yang Feng. 2022{\natexlab{d}}.
\newblock \href {https://doi.org/10.18653/v1/2022.acl-long.467} {Reducing position bias in simultaneous machine translation with length-aware framework}.
\newblock In \emph{Proceedings of the 60th Annual Meeting of the Association for Computational Linguistics (Volume 1: Long Papers)}, pages 6775--6788, Dublin, Ireland. Association for Computational Linguistics.

\bibitem[{Zhang and Feng(2023{\natexlab{a}})}]{DBLP:journals/corr/abs-2303-00257}
Shaolei Zhang and Yang Feng. 2023{\natexlab{a}}.
\newblock \href {https://openreview.net/forum?id=9y0HFvaAYD6} {Hidden markov transformer for simultaneous machine translation}.
\newblock In \emph{The Eleventh International Conference on Learning Representations}.

\bibitem[{Zhang and Feng(2023{\natexlab{b}})}]{zhang2023unified}
Shaolei Zhang and Yang Feng. 2023{\natexlab{b}}.
\newblock \href {https://proceedings.neurips.cc/paper_files/paper/2023/file/8df705957a5262de3cb37ba9f1fb96f3-Paper-Conference.pdf} {Unified segment-to-segment framework for simultaneous sequence generation}.
\newblock In \emph{Advances in Neural Information Processing Systems}, volume~36, pages 45235--45258. Curran Associates, Inc.

\bibitem[{Zhang et~al.(2022)Zhang, Guo, and Feng}]{zhang-etal-2022-wait}
Shaolei Zhang, Shoutao Guo, and Yang Feng. 2022.
\newblock \href {https://doi.org/10.18653/v1/2022.findings-emnlp.166} {Wait-info policy: Balancing source and target at information level for simultaneous machine translation}.
\newblock In \emph{Findings of the Association for Computational Linguistics: EMNLP 2022}, pages 2249--2263, Abu Dhabi, United Arab Emirates. Association for Computational Linguistics.

\bibitem[{Zhao et~al.(2023)Zhao, Fan, Luo, Jing, Wang, Zeng, and Huang}]{zhao-etal-2023-adaptive}
Libo Zhao, Kai Fan, Wei Luo, Wu~Jing, Shushu Wang, Ziqian Zeng, and Zhongqiang Huang. 2023.
\newblock \href {https://doi.org/10.18653/v1/2023.emnlp-main.293} {Adaptive policy with wait-k model for simultaneous translation}.
\newblock In \emph{Proceedings of the 2023 Conference on Empirical Methods in Natural Language Processing}, pages 4816--4832, Singapore. Association for Computational Linguistics.

\bibitem[{Zheng et~al.(2020)Zheng, Liu, Zheng, Ma, Liu, and Huang}]{DBLP:conf/acl/ZhengLZMLH20}
Baigong Zheng, Kaibo Liu, Renjie Zheng, Mingbo Ma, Hairong Liu, and Liang Huang. 2020.
\newblock \href {https://doi.org/10.18653/v1/2020.acl-main.254} {Simultaneous translation policies: From fixed to adaptive}.
\newblock In \emph{Proceedings of the 58th Annual Meeting of the Association for Computational Linguistics, {ACL} 2020, Online, July 5-10, 2020}, pages 2847--2853. Association for Computational Linguistics.

\end{thebibliography}

\appendix

\section{Explanation of Latency Constraint}
\label{sec:latency_constraint}
In the Streaming Self-Attention (SSA) mechanism, the model obtains the attention probability allocated to the source prefixes by predicting $p_{i,j}$. Subsequently, the model accumulates $p_{i,j}$  to determine the total attention on the input source tokens. The model utilizes $\sum_{j=1}^{m} p_{i,j}$ to assess the sufficiency of the source information, thereby acquiring the translation policy. The model generates target tokens when it deems the source information sufficient. To ensure that the model generates translations tightly after reading the necessary source information, we introduce the latency constraint that encourages the model to utilize as few source tokens as possible to generate translations.

\begin{figure*}[t]
\centering
\subfigure[Cost Matrix]{
\includegraphics[width=2.07in]{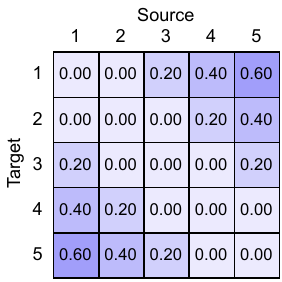}
}\hspace{0.3cm}
\subfigure[Attention Allocation Matrix]{
\includegraphics[width=2.23in]{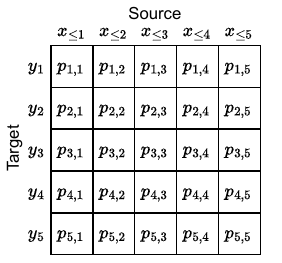}
}

\caption{The illustration of cost matrix and attention allocation matrix. In this diagram, $I$ and $J$ are both set to $5$, and $\epsilon$ is set to $1$.}
\label{matrics}
\end{figure*}

According to \citet{ITST}, the alignments conducive to latency tend to concentrate near the diagonal between source and target sentences. Therefore, we introduce $\mathcal{L}_{lat}$ to encourage SSA to pay more attention to the source prefixes near the diagonal. The intuitive illustration of the cost matrix $\mathbf{C}=(C_{i,j})_{I\times J}$ and the attention allocation matrix $\mathbf{p}=(p_{i,j})_{I\times J}$ is shown in Figure \ref{matrics}.

\section{System Settings}
\label{system_setting}
The system settings on three translation tasks are shown in Table \ref{tab:hyperparameter}.

Our method is based on fine-tuning full-sentence translation model. During the full-sentence training stage, our model adopts the traditional decoder layer \citep{DBLP:conf/nips/VaswaniSPUJGKP17} and is trained with cross-entropy loss. Once we obtain a well-performing translation model, we then add policy-specific parameters and train the model according to our proposed strategy in Section \ref{training_method}.

For more detailed implementation issues, please refer to our code.

\section{Detailed Results}
In addition to the performance comparison in Figure \ref{main_res}, we also present the numerical results for our method. Table \ref{envi}, \ref{enro}, and \ref{deen} respectively describe the performance of DST on  IWSLT15 En$\rightarrow$Vi, WMT16 En$\rightarrow$Ro, and WMT15 De$\rightarrow$En translation tasks. Each task is evaluated using latency measured by AL \citep{DBLP:conf/acl/MaHXZLZZHLLWW19} and translation quality measured by BLEU \citep{BLEU}.

\begin{table*}[t]
\centering
\caption{Hyperparameters of DST.}

\begin{tabular}{c|c|ccc}
\toprule
\multicolumn{2}{c|}{\textbf{Hyperparameters}} & IWSLT15 En$\rightarrow$Vi & WMT16 En$\rightarrow$Ro & WMT15 De$\rightarrow$En  \\
\bottomrule
\multirow{4}{*}{Decoder} & decoder\_layers & 16 & 16 & 16 \\
& decoder\_embed\_dim & 512 & 512 & 512 \\
& decoder\_ffn\_embed\_dim & 1024 & 2048 & 2048 \\
& decoder\_attention\_heads & 4 & 8 & 8 \\
\midrule
\multirow{2}{*}{Loss} & $\epsilon$ & 1 & 1 & 1 \\
& $T$ & 2k & 4k & 30k \\

\midrule

\multirow{11}{*}{Training} & dropout                 & 0.1       & 0.3           &0.3                \\
 &optimizer               & adam      & adam          &adam                \\
&adam\_$\beta$          & (0.9, 0.98) & (0.9, 0.98)   & (0.9, 0.98)     \\
&clip\_norm               & 0         & 0             &0                \\
&lr                      & 5e-4      & 5e-4          & 5e-4                \\
& lr\_scheduler        & inverse\_sqrt  & inverse\_sqrt  & inverse\_sqrt     \\
&warmup\_updates          & 4000      & 4000          & 4000            \\
&warmup\_init\_lr          & 1e-7      & 1e-7          & 1e-7                \\
& weight\_decay            & 0.0    & 0.0        & 0.0        \\
& label\_smoothing         & 0.1       & 0.1           & 0.1                \\
&max\_tokens              & 16000     &  8192$\times$4 & 8192$\times$4 \\

\bottomrule

\end{tabular}
\label{tab:hyperparameter}
\end{table*}

\begin{table}[tp]
\centering
\begin{tabular}{p{2cm}<{\centering} p{2cm}<{\centering} p{2cm}<{\centering}} 
\toprule[1.5pt]
    $\delta_{infer}$      & AL  & BLEU   \\
\cmidrule(lr){1-1} \cmidrule(lr){2-3}
       0.30   &3.89     &28.56   \\
    0.40       &5.32     &29.28  \\
    0.45       &6.37     &29.20  \\
    0.50      &7.70     &29.28  \\
    0.60      &9.75     &29.40 \\
\midrule[1pt]
\end{tabular}
\caption{Numerical results on IWSLT15 En$\rightarrow$Vi.}
\label{envi}
\end{table}

\begin{table}[]
\centering
\begin{tabular}{p{2cm}<{\centering} p{2cm}<{\centering} p{2cm}<{\centering}} 
\toprule[1.5pt]
    $\delta_{infer}$      & AL  & BLEU   \\
\cmidrule(lr){1-1} \cmidrule(lr){2-3}
    0.25   &0.60     &21.01   \\
    0.30       &3.60     &29.80  \\
    0.40       &4.09     &30.73  \\
    0.50      &5.77     &31.63  \\
    0.55     &7.01     &32.62 \\
    0.65     &9.52     &33.08\\
\midrule[1pt]
\end{tabular}
\caption{Numerical results on WMT16 En$\rightarrow$Ro.}
\label{enro}
\end{table}

\begin{table}[]
\centering
\begin{tabular}{p{2cm}<{\centering} p{2cm}<{\centering} p{2cm}<{\centering}} 
\toprule[1.5pt]
    $\delta_{infer}$      & AL  & BLEU   \\
\cmidrule(lr){1-1} \cmidrule(lr){2-3}
    0.20   &3.22     &28.57   \\
    0.30       &4.72     &30.55  \\
    0.40       &5.49     &30.89  \\
    0.50      &7.09     &31.85  \\
    0.60     &9.40     &32.22 \\
\midrule[1pt]
\end{tabular}
\caption{Numerical results on WMT15 De$\rightarrow$En.}
\label{deen}
\end{table}

\end{document}